\def\year{2019}\relax
\documentclass[letterpaper]{article} 
\usepackage{aaai19}  
\usepackage{times}  
\usepackage{helvet} 
\usepackage{courier}  
\usepackage[hyphens]{url}  
\usepackage{graphicx} 
\def\UrlFont{\rm}  
\usepackage{graphicx}  
\usepackage{todonotes}

\usepackage{subcaption}
\usepackage{enumitem}
\usepackage{booktabs}
\usepackage{pbox}
\usepackage{amssymb}
\usepackage{comment}
\usepackage{dblfloatfix}
\title{Designing Evaluations of Machine Learning Models for Subjective Inference: The Case of Sentence Toxicity}
\author{Agathe Balayn, Alessandro Bozzon
\\ \Large 
Delft University of Technology, the Netherlands\\ 
\{a.m.a.balayn, a.bozzon\}@tudelft.nl 
}
 
\begin{document}

\maketitle

\begin{abstract}
Machine Learning (ML) is increasingly applied in real-life scenarios, raising concerns about bias in automatic decision making. We focus on bias as a notion of {opinion exclusion}, that stems from the direct application of traditional ML pipelines to infer subjective properties. 
We argue that such ML systems should be evaluated with subjectivity and bias in mind. Considering the lack of evaluation standards yet to create evaluation benchmarks, we propose an initial list of specifications to define prior to creating evaluation datasets, in order to later accurately evaluate the biases.
With the example of a sentence toxicity inference system, we illustrate how the specifications support the analysis of biases related to subjectivity. We highlight difficulties in instantiating these specifications and list future work for the crowdsourcing community to help the creation of appropriate evaluation datasets.
\end{abstract}

\section{Introduction}
Machine Learning (ML) is increasingly used in real-life applications. 
However, for the sake of simplification, 
researchers and engineers are not interested in the specific requirements of the applications, but target the development of ML models with generalization abilities. Hence the systems might not be adapted to their end-users (EUs). 
Specifically, we notice a shift of focus from using objective labels (e.g. digit recognition ~\cite{lecun1990handwritten}) towards subjective labels (e.g. rating of image aesthetics~\cite{bianco2016predicting}) for classification or regression problems. ML pipelines are often not designed with subjectivity in mind, which results in creating systems whose outputs carry biases. For example, the opinion of certain individual EUs are ignored, leading to a negative user experience for them. 
Consistently ignoring the opinion of the same group of EUs, the minorities, might reinforce filter bubble effects, an emergent  danger for societies~\cite{bozdag2015breaking}.

The development process of real-life applications 
merits being investigated in more details to counter these issues. We consider current processes as vicious circles: developers' goals are not adapted to the application and hence the evaluation is not either - which prevent from identifying the lack of adaptation but reinforces it. With proper evaluation frameworks
, we could audit and build more adapted systems. Thus we investigate here ML evaluation settings, that we see as a combination of an evaluation dataset - for which the ground truth is collected via crowdsourcing - and evaluation metrics to compute on the dataset and the predicted labels of the ML model.
We focus on subjective classification tasks because it is rarely studied but highly relevant to the crowdsourcing community as crowdsourcing is at the start of their development process (dataset collection). 

No standard exists for the evaluation of these applications. 
Hence we propose a brief formalisation and establish initial specifications that the evaluation settings should consider and aim to follow in order to better account for subjectivity. We expose through the analysis of one use-case the difficulties to establish instantiations of these specifications and their resultant requirements, and highlight directions of future work
. We illustrate how current ML practices are limited 
and show that our specifications enable deeper and more accurate understanding of ML systems. 
Finally, we focus on the crowdsourced collection of data labels. We investigate whether the gaps between ML practices and requirements can be filled by state of the art crowdsourcing methods or require additional research.
We work on the use-case of inferring sentence toxicity because it is a subjective property. There are both a fair amount of literature in Social Sciences to understand it, and a large amount of Computer Science literature to address the problem (and a large benchmark dataset) which enables to analyse ML practices.
Besides, it is {urgent to develop efficient automatic toxic speech detection techniques}~\cite{tsesis2001hate} to help human moderators.



\section{Background}

ML systems for subjective applications and their evaluation are rarely examined. Related research is spread in two main communities. 
The crowdsourcing community focuses on the collection of subjective labels, in relation with aggregation bias creating a loss of information~\cite{balayncharacterising}. The ML community designs ML algorithms 
which mostly 
ignore the subjectivity. 
None of them address the creation of evaluation datasets for subjective inference tasks.

\subsection{Crowdsourcing Community: Quality, Biases}
Although not for the purpose of creating evaluation datasets, the crowdsourcing community investigates how to collect quality labels from data samples. 
The quality 
depends on three concepts~\cite{dumitrache2015achieving} 
from which disagreement arises and where subjectivity is involved
: expertise and performance of the crowd workers (CWs), design of the task 
and ambiguity of samples. Methods for ensuring labels' quality investigate these concepts.

General methods (e.g. task design, user interface
~\cite{alonso2011design}
) which mostly consider quality as {data without disagreement} are not adapted to subjective cases which tolerate disagreement due to CWs' subjectivity. Methods for subjective tasks leverage disagreement but mostly aim at finding a {unique ground truth} (GT) often available through experts. 
Collaborative approaches of discussion between CWs to refine the task labels
~\cite{chang2017revolt,drapeau2016microtalk} imply agreement on a unique GT and influence individual subjectivities.
CWs' quality is assessed to weigh out poor quality CWs 
by exploiting task features without implication for the subjectivity (e.g. gold standard questions
, consistency tests
, open-ended questions, 
user monitoring
~\cite{hossfeld2011quantification,redi2013crowdsourcing
}), or by taking into account the subjectivity 
based on CWs' labels
~\cite{ribeiro2011crowdsourcing} 
or on CWs' labels and few diverse expert labels~\cite{speck2011comparative}. 
The CrowdTruth framework~\cite{aroyo2013crowd} proposes metrics which account for the disagreement to assess the CWs, the samples' clarity and the labels' relations
.
To avoid propagating mistakes
, labels of several CWs are {aggregated into a unique GT} (e.g. by majority-voting -MV or probabilistic approach~\cite{raykar2010learning
}), what loses information from diverse valid labels. 
Groups of CWs with similar label trends can be identified by clustering to discover CWs 
having specific interpretations of the tasks, but this is not yet scalable
~\cite{kairam2016parting}.

Research rarely relates CWs' biases to labels' subjectivity and ML biases. 
CWs' biases are related to {diversity}, composed of diversity of identity, skills and political investment, the first one being the most influential for the quality of crowdsourcing~\cite{brabham2008crowdsourcing}.
{Individual biases from the interpretation} of a scale~\cite{snow2008cheap} are removed by {recalibrating labels} with the aim of having a unique interpretation of the task. 
{Demographic biases} 
from the CWs' pool
result in various labels per sample~\cite{ghadiyaram2016massive}. \cite{barbosa2019rehumanized} propose an optimized task allocation mechanism which prepares a balanced, debiased crowd to account for relevant CWs' biases due to their backgrounds (e.g. language, expertise) in datasets. 

\subsection{ML Community: subjectivity, unfairness}

Research on the inference of subjective properties {{ignores subjectivity by removing disagreement}} through label aggregation~\cite{bianco2016predicting
} 
removing the data with the most disagreement
~\cite{jamison2015noise
}. 
Soft labels are employed~\cite{sharmanska2016ambiguity
} to account for diversity but do not convey each EU's opinion. 
\cite{alonso2016supersense} exploit CWs' disagreement as an error signal to train ML models 
but not as valid subjectivity. 
Hence subjectivity is not accounted in the outputs and evaluation of the models.
Conversely in {{personalisation, ML models' outputs are adapted to the EUs}} (e.g. response generation
~\cite{covington2016deep
}, recommender systems), and the evaluation is done with usual performance metrics. 



Only \cite{binns2017like} account for the diversity of opinions and initiates the study of dataset biases in subjective inference tasks and the repercussions on ML outputs. They consider that an unbiased (fair) model has equal performance among the categories of population in the dataset (male, female) and show that implicit norms of CWs lead to discrimination in ML models. They train separate models on the opinions of the different categories but mention there is also disagreement within the categories. 
Similarly \cite{reidsma2008exploiting
} account for disagreement 
by training one classifier per CW, which is not scalable. 
Our work is closer to these works.
They are related to ML research on fairness because they consider that not accounting for user biases in the outputs of ML models is unfair and use unfairness metrics to quantify the extent of this repercussion.  

Most works see unfairness as discrimination towards protected categories of population,  
in models which classify people (e.g. whether someone who committed a crime will reoffend) using protected 
and non-protected features
~\cite{gajane2017formalizing}. 
Papers agree on the definition of unfairness~\cite{
binns2018fairness} but differ on 
its formalization within metrics and mitigation approaches.


\section{Evaluation setting: formalisation}
\begin{figure}[h]
  \centering
  \includegraphics[width=0.9\linewidth]{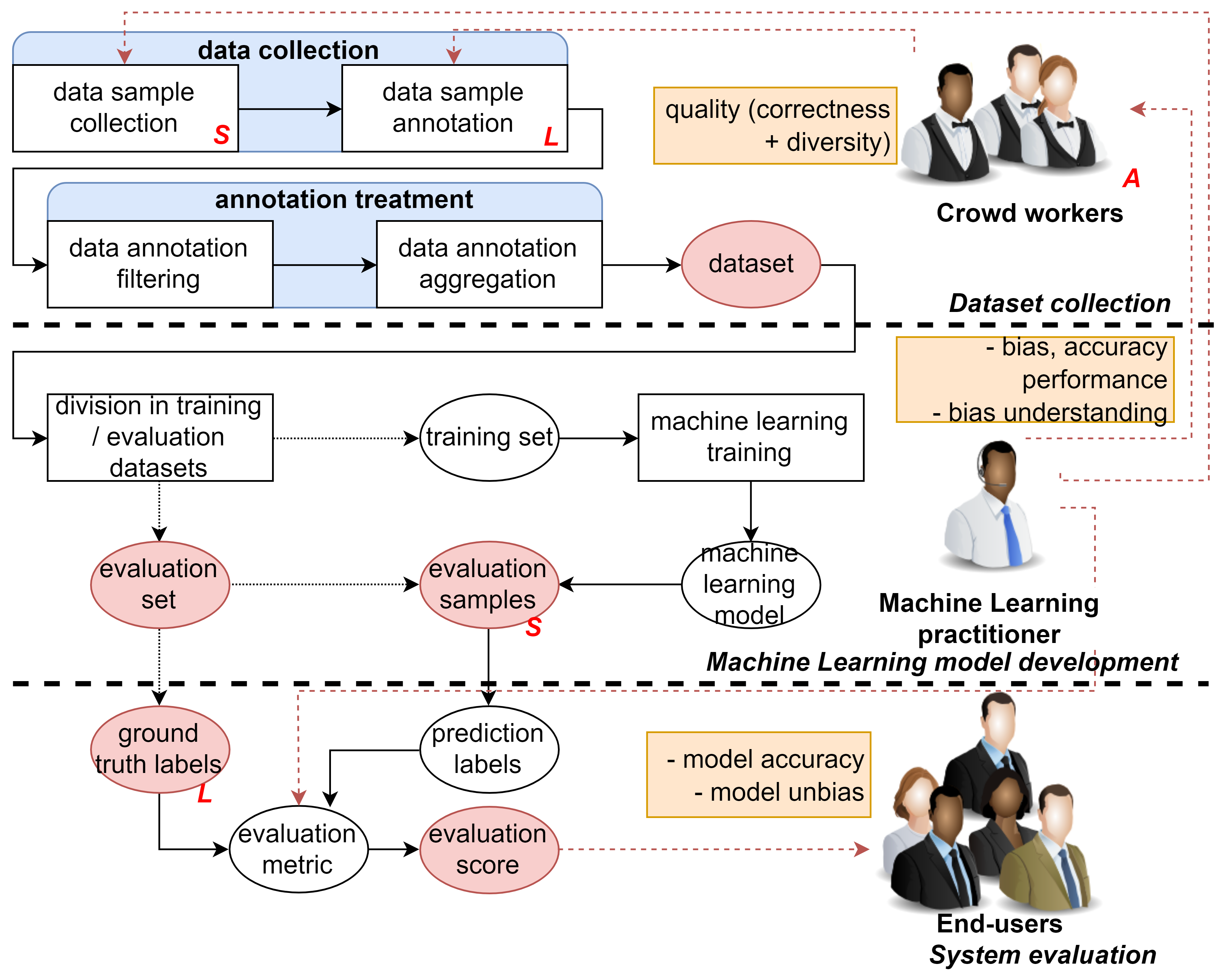}
  \caption{Current evaluation settings for ML models.}
 \label{fig:current_crowdsourcing_steps}
\end{figure}

Fig.~\ref{fig:current_crowdsourcing_steps} summarises the steps involved in the current development process of an ML system, with its stakeholders and their requirements.
The evaluation process of ML models traditionally relies on the computation of a set of metrics on an evaluation dataset and the associated predictions of the model.
The goal of the first steps is to create a dataset $D = (S, L, A)$. It is a tuple of a list $S \in \mathbb{R}^{n\times e}$ of $n$ data samples (with $e$ the encoding dimension of the samples), with $l$ multiple associated labels $L \in \mathbb{Z} ^ l$, representing the opinions per sample of the $f$ diverse users in $A \in \mathbb{R}^{f\times u} $ (with $u$ the number of user features)
. This dataset is split and used for training and evaluating the ML model.
Next we refer to $D$ of the evaluation specifically.
We identify the implications on the elements of $D$ that the expectations of the stakeholders have, and draw a list of specifications that the evaluation setting extracted from $D$ should respect.

\section{Data collection: specifications, implications}
The evaluation 
must account for subjectivity 
to avoid the development of biased systems.
Its specifications 
should address the interconnection between metrics and evaluation datasets, and its influence on the interpretation 
of the bias of 
the system. 
Hence, we reflect on 
a list of specifications to 
define 
before setting up an evaluation process and collecting a dataset. 
We hope this initial list triggers reflections on the needed establishment of specifications, notably by presenting the difficulties in defining instantiations of such specifications for any application, and on the research needed to fulfill the specifications in practice.


\begin{enumerate}[wide, labelwidth=!, labelindent=0pt]
    \item \textit{Choice of bias definition and metric}.
    Bias is context dependent, hence its definition and metrics 
    should be selected through a careful study of the application. 
    ML fairness literature identifies two main types of fairness: group fairness where groups of people for which only the protected attribute differs should be treated equally, individual fairness where similar individuals should be treated equally.
    \item \textit{Definition of user features.}
    The computation of the {similarity metric} for individual fairness relies on  $u_S$ knowledge about individuals in $A\in \mathbb{R}^{f\times u}$. 
    Hence the {$u_P$ protected attributes
    } and $u_S$ similarity features (which might overlap) 
    should be identified before data collection and contained  in $A$ for each individual to evaluate bias.
Additional $u_R$ user features on which to base the representativeness of the dataset should vary in $A$ to compute accurate performance metrics.
To define them and their ranges
, we propose to bring domain knowledge and identify variables influencing end-users (EUs)' opinions (giving $u_P$, $u_S$ and additional features $u_R$, where 
$u = u_P + u_S + u_R$), and to obtain prior knowledge on the EUs. 
The list of user features influencing judgements might be long, hence the complexity of defining $A$ and similarity metrics, and the large amount of individual knowledge required to perform an exhaustive evaluation. 
\item \textit{Definition of a target distribution for $A$.}
The specifications on $A$ should allow the computation of statistically valid metrics.
Hence 
the distribution of $A$ should follow the distribution of or be balanced over EUs' user features depending on the metrics. 
We suggest to target for each sample  
an equal number of EUs with varied user features 
which influence opinions~\cite{schmidt2017survey}
, as a start to foster diversity and representativeness of labels. 

\item \textit{Definition of a target distribution for $S$.}
The distribution of $S$ should be representative of the samples used in the real application, which implies prior knowledge on the application. For instance the source of $S$ should be similar to the real application.
Further specifying $S$ might be easier by identifying characteristics of the samples with a thorough analysis of the application and aiming for them in $S$, than after encoding the raw data into less interpretable features.

\item \textit{Definition of a target distribution for $L$ through $S \times A$.}
$L$ should be representative of the diversity of opinions of the EUs on the diverse samples of the application.
Hence, after having defined the target $A$ and $S$, the distribution of $L$ is specified according to the metric requirement, mostly as an equal amount of labels for each sample in $S$ provided by an equal amount of crowd workers (CWs) in each category in $A$, amounts enabling statistically correct computations.

\item \textit{Collection of $S$, $A$ and $L$}.
We recommend to first collect the samples according to $S$ in order to feed them to the crowdsourcing task. 
Then we need to recruit a pool of CWs according to $A$ and their labels according to $A \times S$.
Several technical difficulties arise.
Since the labels are most often collected via crowdsourcing, the population of CWs should be an accurate proxy for the EUs. It is difficult to identify, recruit and retain 
CWs fulfilling target $A$. Work on the creation of additional crowdsourcing platform plug-ins to identify CWs with certain features and to evaluate CWs' features (for instance through self evaluation which might be biased and unreliable or with psychology tests) 
is needed.
Retaining the CWs to annotate all samples is impossible, and hence it is required to recruit additional CWs with the same specifications 
to replace them in the distributions of $A$ and $A \times S$. 
The cost and time to reach $L$ might be high to achieve statistically correct results. Statistical methods to propagate annotations and crowdsourcing methods to optimize the obtention of target distributions would merit being studied (e.g. \cite{chung2019efficient}). 
Independently from the distributions, labels should be diverse and of high-quality to represent valid opinions - this is 
investigated in the next section.
This implies iterations in collecting $A$ and $L$ since certain CWs and labels might be eliminated for quality reasons.
Furthermore, EUs and samples of the real life applications might change over time, hence the target $D$ might have to be adapted and the collection process repeated.
We foresee the evaluation to be an iterative process where more samples and CWs could be needed to investigate different types of biases after receiving preliminary results. Adding new CWs' attributes to $A$ is difficult since CWs might not be identifiable, hence we recommend to collect more attributes first while considering cost constraints. 



\item \textit{Reporting of the specifications.} 
Due to the difficulties and subjectivity in defining the specifications
, we argue for the need to save metadata about the dataset 
(similarly to \cite{DBLP:journals/corr/abs-1803-09010}) in order to allow reproducibility of the evaluation (and potentially extension), and through this transparency to allow critical interpretation of the bias/fairness scores. Specifically, we suggest to report the chosen definitions and metrics for biases, the target distributions defined previously with the reasoning behind them (e.g. knowledge about EUs and specification on $A$) and the actual distributions employed. Furthermore, scores computed over the whole evaluation set do not enable to understand the causes of bias. Hence we argue that evaluations would gain usefulness if intermediate scores for individuals or groups used in the metrics' computations were reported. 
The composition of these groups (amount of data points in the groups and features) should be reported while releasing the datasets. 
\end{enumerate}


\section{Illustration of the 
specifications}

We conduct experiments on the use-case to demonstrate the reasoning behind the specifications.
Systems for toxicity inference would benefit taking it into account subjectivity~\cite{alm2011subjective}. 
However, ML papers only use traditional performance metrics (e.g. precision, recall) except~\cite{binns2017like}
. 
None checks for the specifications of the datasets. These evaluations do not account for biases rising from the subjectivity~\cite{balayncharacterising}.
We use the Jigsaw dataset~\cite{wulczyn2017ex} 
because it is the largest public one, and it gives access to the annotations of the CWs.
It consists in 100000 Wikipedia comments whose toxicity is rated on a $[-2;2]$ scale~\footnote{-2 toxic: "a very hateful, aggressive, or disrespectful comment that is very likely to make you leave a discussion.", 2 healthy: "a very polite, thoughtful, or helpful contribution that is very likely to make you want to continue a discussion."}. 10 labels per sample 
    were collected on CrowdFlower with information about 3500 
    CWs (gender, language, age group, education). 

\begin{enumerate}[wide, labelwidth=!, labelindent=0pt]
    \item  We define a {system to be biased when its performance is unequal across its end-users (EUs)} (opinions would not be equally accounted for among EUs, e.g. the system could return only the opinions of the majority
). 
We use two metrics, one focusing on protected attributes and one on groups of similarly thinking crowd workers (CWs) without assumption on attributes. Both metrics rely on {{grouping sample-label pairs}} in the dataset with respect to CWs' category who provided them. This way metrics computed with datasets of similar groupings are comparable. For the first one, CWs' category is defined by the protected attributes
. For the second one, categories are attributed to CWs based on their Average Disagreement Rate (ADR) with the majority-vote (MV) (average number of labels of a CW which differ from the MV) 
because the MV represents the common judgement ; and that enables to see whether CWs all follow the same line of thoughts. 
After grouping, the model {performance for each CW and the mean of these values within each group} are computed. Finally, the metric produces a pair of values (disp., perf.): the {(1 - the standard deviation) and mean across groups} to respectively quantify bias and performance. 
    The dataset range of the grouping criteria 
    can have a large effect on the evaluation. 
   Hence we report the range used or compute the metric with the range as parameter. 
    \item We use Psychology literature as the expert to understand the knowledge on people to account for.
    There, toxicity judgements depend on three types of factors~\cite{guberman2017challenges,downs2012predicting,williams2016racial}. 
    People's inner characteristics: 
    nature of the judge (gender, age, ethnicity, education) 
    and on other traits and experiences. 
     People's perception of a sentence characteristics: categories of hate and targets
     and syntactic and semantic properties. 
     Sentence context: author, receiver
    , public or private speech and community in which the sentence is announced. 
    The age, gender and education are recurring and hence these are the protected attributes we select. 

\item No prior target distribution was defined for $A$. The number of labels between categories of CWs (for each protected attribute or their combinations, see Fig.~\ref{fig:CW_distribution}) is highly unbalanced with predominant groups being the typical CWs while others are difficult to recruit (e.g. 1862 CWs are between 18 and 30 years old, whereas only 30 are more than 60 years old
). This affects the next steps since the metric computations are not statistically correct for the rarest groups.
\item There is no real application at hand, hence no conclusion can be made on the sample distribution. The data were crawled from Wikipedia conversations, possibly representative of Web discussions. $S$ seems fairly representative because most samples receive neutral judgements (and not toxic labels), what is expected from Wikipedia (Fig.~\ref{fig:CW_distribution1}).
\item Due to the low number of CWs for certain categories of attributes, the number of labels for each sample coming from these different categories is unbalanced (having only 10 labels per sample is too few compared to the number of combinations of user features' categories possible here).
\item The dataset was collected without subjectivity in mind: no attention was given to CWs' subjectivity but only to labels' correctness seen as having a unique valid ground truth created via the aggregation of multiple CWs' inputs.
\item We show that reporting intermediate results hints on causes of biases and enables to understand limitations of the evaluation dataset. For this, 
we create 3 systems with various expected bias behaviours using a simple classifier (Logistic Regression). 
Model 1 is a biased model trained on samples and the MV. Model 2 is trained on 
samples, user features and corresponding opinions, and is thus expected to be less biased. Model 3 is an hypothetical accurate and unbiased model. 
We {{display the group performance}} used to compute the metrics to identify the types of people treated unfairly.
~\footnote{The tool and metric are available in the AIF360 IBM toolkit~\cite{aif360-oct-2018} with examples of models' evaluation.} 

\begin{figure}[tb!]
\centering
\begin{subfigure}[b]{0.44\linewidth}
		\includegraphics[width=1\textwidth]{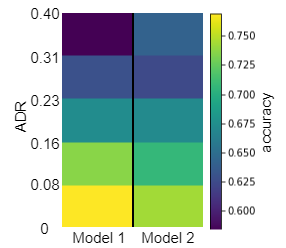}
  \caption{CWs grouped on ADR.}
  \label{fig:unfairness_visualization}
\end{subfigure}
  \begin{subfigure}[b]{0.53\linewidth}
		\includegraphics[width=1\textwidth]{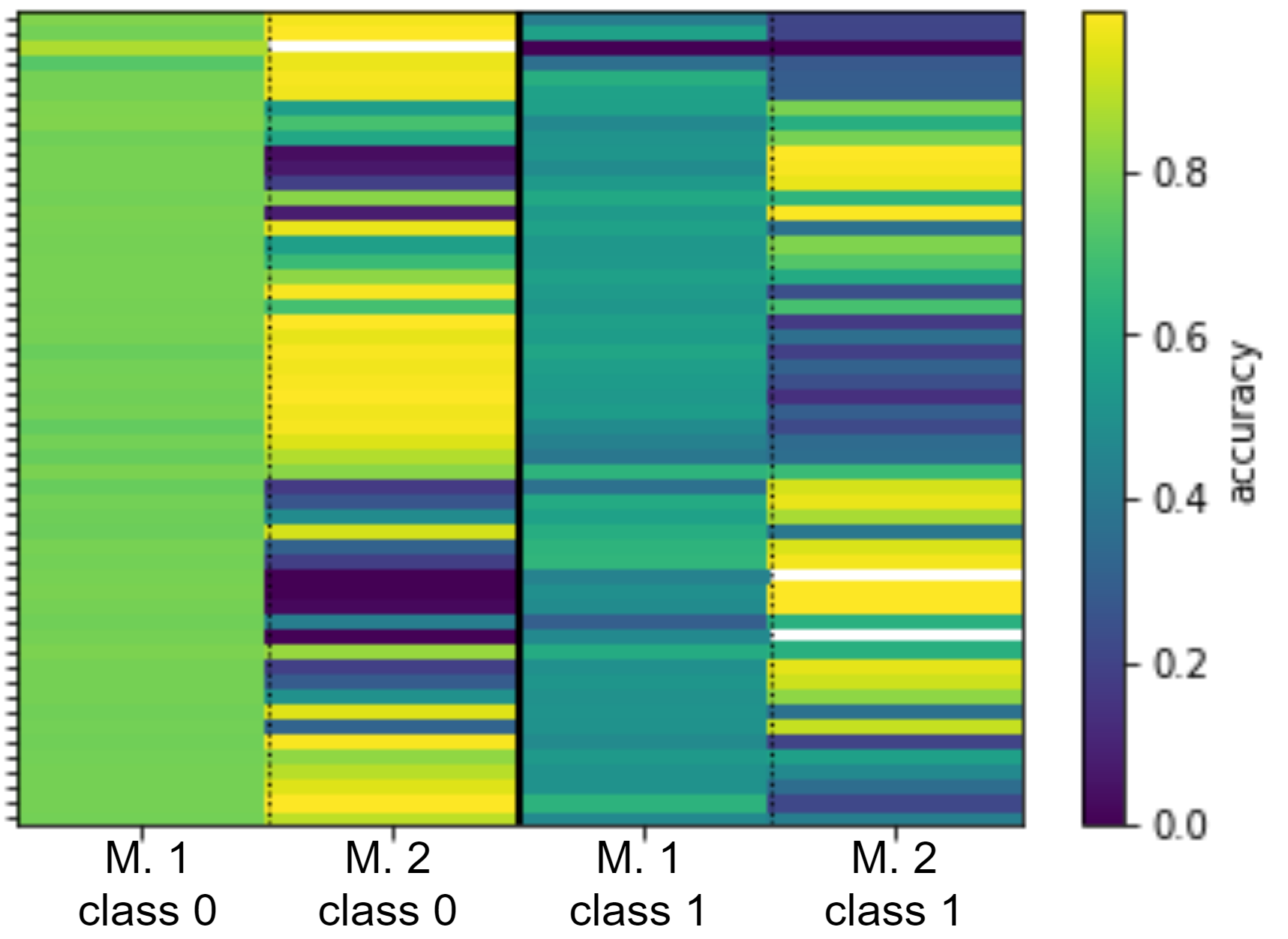}
  \caption{Groups of protected attributes.}
  \label{fig:unfairness_visualization2}
\end{subfigure}
\caption{Bias exploration tool: ML performance}
\vspace{-6pt}
\end{figure}
In Appendix, we report the results of the metrics.  
Fig.~\ref{fig:unfairness_visualization} is an example of group performance visualisation for the second metric. It correctly shows that model 1 is more biased than model 2 since its performance across groups (rows) are more disparate. The types of EUs which lead to bias are highlighted: model 1 is inaccurate at predicting opinions of EUs who disagree with the MV the most, this is better in model 2 but not fully solved ; while performance for EUs who mostly agree are much higher. It shows that if the evaluation dataset was only constituted of CWs of the top group, both models would appear unbiased leading to wrong interpretations. 
Fig.~\ref{fig:unfairness_visualization2} shows that model 2 is more biased than model 1 regarding protected attributes
, and highlights user categories that need more training data. If the evaluation dataset did not contain data points for all the categories
, model 2 could appear unbiased, producing a wrong evaluation.
If the characteristics of the evaluation dataset (e.g. grouping criteria, distributions) were specified as meta-data, it could avoid missinterpretations, especially if the dataset is used for applications different from the initial one. 
\end{enumerate}

\section{Crowdsourcing treatment of annotations}

We now identify the gaps between the ML practices with regard to the treatment of subjective crowdsourced data annotation (filtering and aggregation) and the expectations on $D$ (correct and diverse annotations representative of the EUs' opinions). We assess whether the state-of-the-art in the crowdsourcing domain is adapted to the requirements of subjective labeling tasks, and draw a list of recommendations for future crowdsourcing and ML research.

ML researchers use crowdsourcing to collect toxicity datasets, because unlike tasks such as 
subjectivity labeling~\cite{hsueh2009data}, it is hard to define experts for toxicity labels. Due to potential spammers and mistakes from the CWs, 2 to 5 CWs provide annotations for the same samples, aggregated into unique labels (e.g. by majority voting MV) in the hope of obtaining valid labels. This is not adapted to the requirements for subjective classification tasks where 
different labels might be valid for different people. 
Hence we investigate where {incorrect labels} might come from and how to eliminate them, while keeping the valid {diversity} (trade-off correctness/diversity).
Related work outlines the degrees of freedom of crowdsourcing to consider for improving label quality. 
a) Clear task design methods aim at removing ambiguity, which has consequences on the correctness. 
b) Recruitment of CWs, c) online filtering of CWs' labels by CWs' expertise, pre-training or CWs' platform's quality scores 
and d) post-treatment 
of labels impact correctness and diversity.
Methods for {a), b), c) are not specific to objective data and thus can be used to improve correctness of the dataset} (refinement in light of knowledge on subjectivity could improve them).
Conversely, most {techniques to post-filter CWs} (d) are not adapted because they assume a unique GT or the existence of experts or are applicable only to tasks where all CWs labeled the same samples, except {CrowdTruth}~\cite{dumitrache2018metrics} which {leverages disagreement}. We assess whether they are used by ML practitioners and usable in subjective tasks. 


\subsection{Design of tasks for annotation collection}
Crowdsourcing research aims at making the task as clear as possible in order to facilitate its understanding for the CWs to be fast and accurate at providing labels. However this is not followed in current ML datasets. Instructions are not precise: terms ("toxicity") and scale of the question are not explained, no example is given, no guidelines are specified for ambiguous cases where CWs could have different interpretations.
Not only the ML community should follow the guidelines from the crowdsourcing domain, but it should also analyze the dataset itself to identify and exemplify complex samples' desired labels.
Besides, the design could be adapted from the Psychology experiments. 
There, researchers collect judgements over toxic speech by using questionnaires consisting of multiple questions over a sample, whose answers are averaged. This enables to obtain high-quality data which reflect the CWs' unambiguous opinions. 
Moreover, they use different scales to rate the propositions that could be used in ML not to confuse CWs with multiple potential interpretation of the questions.
Similarly to the crowdsourcing triangle (\cite{dumitrache2015achieving}), Psychology details three factors which influence judgements (sample characteristics, context and judge), subjectivity coming from the judge and characteristics' interpretations. Hence ambiguity (and not subjectivity) might come from unclear context, what ML practitioners should strive to disambiguate along the variables from Psychology (author, receiver, etc.) instantiated within scenarios for Psychology experiments. However no conversation context is present in toxicity datasets.
Tab.~\ref{tab:example_crowdsourcing} gives examples of ambiguous sentences which would be clear if the task design was more detailed, according to previous suggestions.

\begin{table*}[tp]
	\footnotesize
	\centering
	\caption{Example sentences, judgements and potential causes of disagreement.}\label{tab:example_crowdsourcing}
	\begin{tabular}{@{}lcc@{}}
		\textbf{sentence} & \textbf{annotations} & \textbf{causes of disagreement}  \\  \midrule[2pt] 
	\pbox{0.45\textwidth}{\textit{Everywhere, you were also very disruptive as well.}} & \pbox{0.15\textwidth}{-2(1),-1(2),0(5),1(2)} & \pbox{0.35\textwidth}{No context (truth or criticism).}		\\ \noalign{\smallskip}\hline\noalign{\smallskip}
		\pbox{0.45\textwidth}{\textit{Shush sweetie, the adults are talking.} } &  \pbox{0.15\textwidth}{-1(5),0(3),1(2)}
		&  	\pbox{0.35\textwidth}{Lack of context about the actors of the discussion. 
		} \\ \midrule[2pt]
		\pbox{0.45\textwidth}{\textit{I hereby wish to thank you for your continuous efforts in protecting our templates from those gutless vandals [...]
		}}
		&  \pbox{0.15\textwidth}{-1(4),0(2),1(4)}
		&   \pbox{0.35\textwidth}{Lack of instruction to deal with two opposite sentiments in a sentence. 
		} \\ \noalign{\smallskip} \hline \noalign{\smallskip}
		\pbox{0.45\textwidth}{\textit{The transition between the first two paragraphs is horrible. [...]
		the content in para 1 does not appear related to para 2.}} &  \pbox{0.15\textwidth}{-1(3),0(1),1(6)} &  	\pbox{0.35\textwidth}{Lack of instruction (constructive criticism but negative vocabulary).} \\ \midrule[2pt]
			\pbox{0.45\textwidth}{\textit{I already asked. He pretty much told me to stick my head in a bucket of lava.}} & \pbox{0.15\textwidth}{-1(6),0(3),1(1)}
		& 	\pbox{0.35\textwidth}{CW's subjective perception of the toxicity of the sentence.} \\ \noalign{\smallskip}   \hline \noalign{\smallskip}
		\pbox{0.45\textwidth}{\textit{What the hell is wrong with this thing ? why are my changes not showing?}} & \pbox{0.15\textwidth}{-2(1),-1(5), 0(3),1(1)}
		& 	\pbox{0.35\textwidth}{Lack of instructions for sentences without target, or CW's subjectivity.} \\ \hline
	\end{tabular}
\end{table*}




\subsection{Annotation filtering methods}

Concerning filtering of low-quality CWs, crowdsourcing proposes techniques 
independent from the subjectivity. In the Jigsaw dataset, CWs are filtered using 10 golden questions and consistency tests over their answers. Other crowdsourcing methods could also be used, along the multiple-question questionnaires from Psychology which eliminate CWs' mistakes without aggregation. 
These methods are not enough since mistakes are still present, hence the need for post-processing methods.
The analysis of the dataset outlines three kinds of CWs whose labels differ from the MV in subjective tasks: {spammers}
, {CWs who make infrequent mistakes} and {CWs whose interpretation of samples 
differs from the MV}. 
To obtain a dataset of high quality without eliminating diversity, we need to remove the first two types of CWs while making sure not to remove the third type.

\subsubsection{Experiments}
To test whether CrowdTruth enables to identify low-quality labels, we apply it to the dataset and check its validity. It takes as input a set of labels and CWs for several samples. It outputs Unit (UQS), Worker (WQS), and Annotation Quality Scores (AQS), that we compare to a manual evaluation of the lowest and highest quality CWs and samples.
For that, we sample 60 CWs whose WQS scores on binary labels are very low or very high and compute a quality score per CW by averaging the number of their labels considered correct. Then we compute the mean-squared error (MSE) between these scores and the WQS. Low error means CrowdTruth identifies accurately correct and incorrect labels. 
We also sample 100 sentences from the whole range of UQS
, and give binary ambiguity scores depending on whether the sentence 
is subject to multiple interpretations. Because our scores are binary but the UQS are continuous
, we compute the Area Under the Receiver Operating Characteristic Curve (AUROC). 
The higher the result, the more accurately the framework identifies ambiguity. 
We apply the framework to 4 set-ups with 4 scales obtained by 4 label aggregations (set-up 1: -2, -1 toxic and 0, 1, 2 non-toxic ; set-up 2: -2, -1, 0 toxic and 1, 2 non-toxic ; set-up 3: -2, -1 toxic, 0 neutral, and 1, 2 non-toxic ; set-up 4: original labels) to compare the validity of CrowdTruth on 4 scenarios for which labels' quality differs. 
The MSE and the AUROC are computed on the same data samples for each set-up.

\subsubsection{Results and Discussion}
\begin{figure}[tp]
	\centering
	\begin{subfigure}[b]{0.61\linewidth}
		\includegraphics[width=\textwidth]{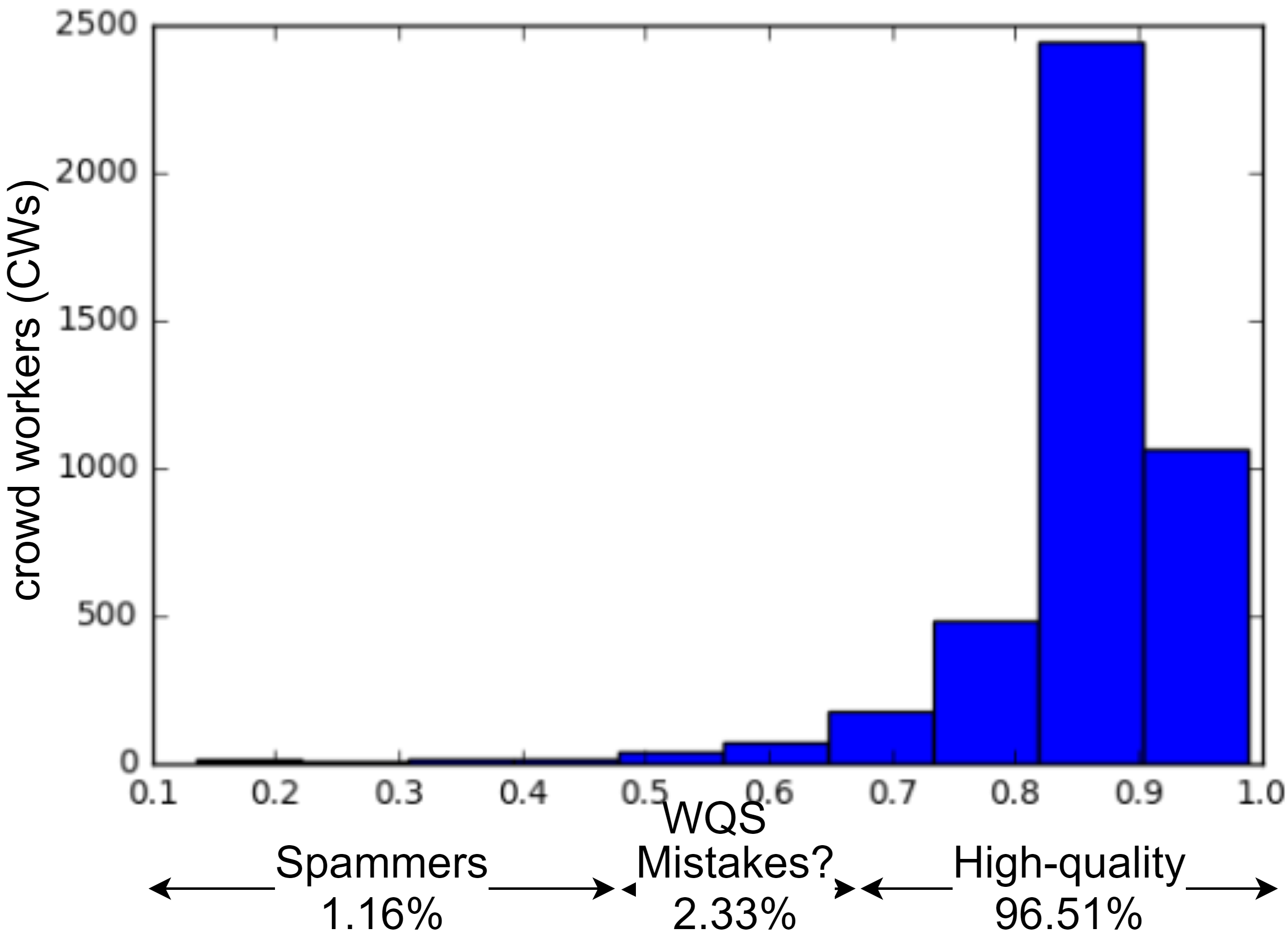}
		\caption{CrowdTruth WQS and types of CWs.}
		\label{fig:WQS_details}
	\end{subfigure}
	\begin{subfigure}[b]{0.34\linewidth}
	\includegraphics[width=\textwidth]{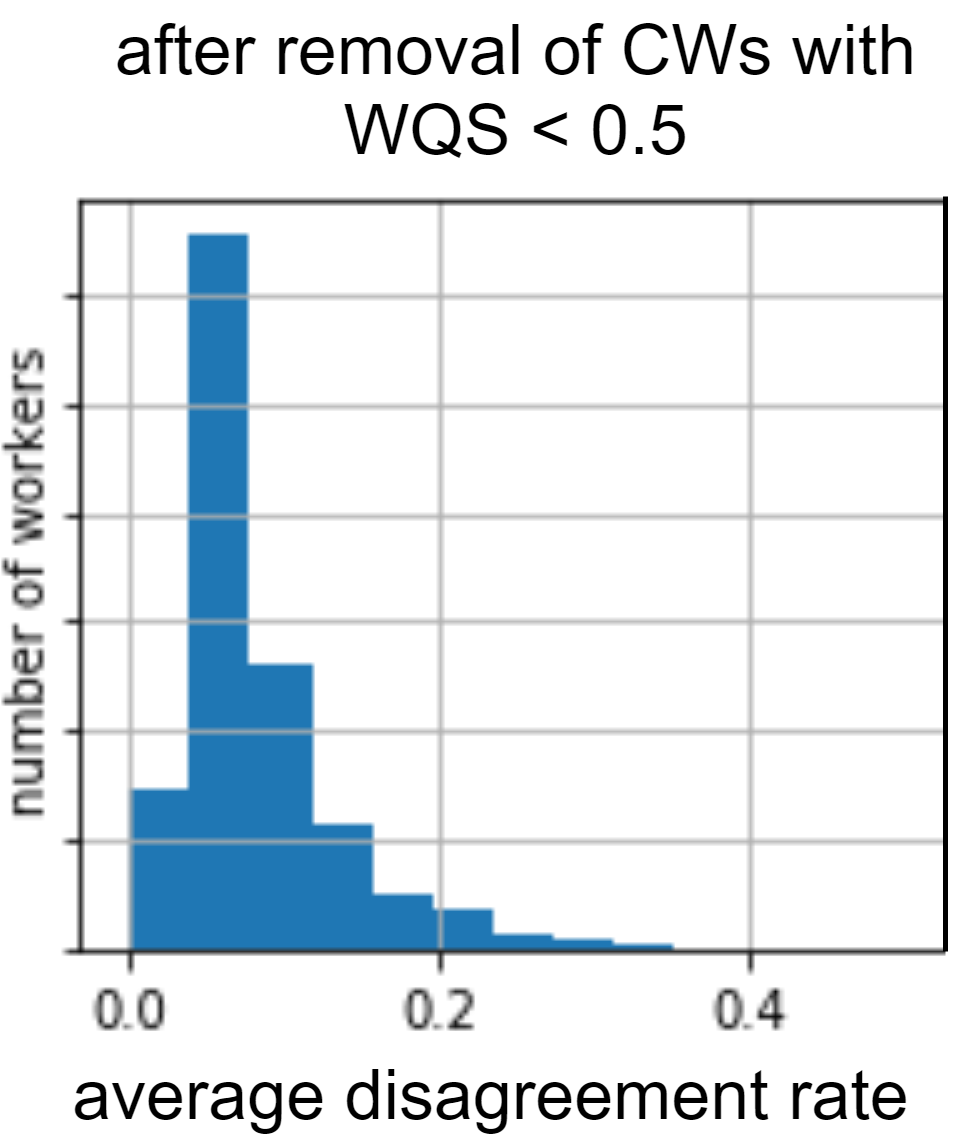}
	\caption{ADR distribution.}\label{fig:ADR_crowd}
    \end{subfigure}
    \caption{CWs' quality and disagreement (set-up 1).}
    \vspace{-12pt}
\end{figure}

We obtain the following {MSE and AUROC 
scores in order of the set-ups: [0.0103, 0.1826, 0.2679, 0.3133], [0.9452, 0.4792, 0.8607, 0.7906]}.
The scores 
are closer to our judgement of the CWs' labels for set-up 1 than the other set-ups. This was expected since set-up 1's labels make the most sense. 
The manual exploration of the samples and CWs supports this interpretation.
For set-up 1, Crowdtruth attributes a high quality to most sentences because the labels are easily interpretable so there are less mistakes: low-UQS samples are sentences in foreign languages and sentences with ambiguous interpretation while high-UQS samples are long constructive comments ; low-WQS CWs provide random labels while high-WQS CWs have mostly valid labels.
For the other set-ups, the scores are concentrated on lower-quality values, 
because there are more errors since the labels are more complex to understand and there are more possible perceptions of the same labels. 
The AQS for set-up 1 are high (non-toxic being the clearest label) and decrease for set-up 2 to 4. 
This trend supports our explanations on the clarity of the labels. 
This indicates that {CrowdTruth results are relevant for subjective crowdsourcing tasks}. 
An analysis of low-WQS CWs shows that CrowdTruth provides sound results with binary labels to eliminate spammers, the WQS threshold needs to be selected by manual analysis of the data. However {it does not differentiate between CWs' occasional mistakes and CWs' valid judgements different from the MV}. These points remain to be investigated in the future, 
but {CrowdTruth is satisfactory for now} since it only involves 2.33\% of CWs (Fig.~\ref{fig:WQS_details}).

We assume that valid subjectivities manifest in the existence of various valid labels on one same sample, and quantify disagreement in the CrowdTruth-processed labels to show their existence. We measure it through the ADR.
We plot the distribution of ADR on different ranges of WQS (example Fig.~\ref{fig:ADR_crowd}). 
After removing spammers, only 10.5\% of the CWs 
always agree with the MV, 51\% disagree around 15\% of the time and 4.5\% disagree at least 20\% of the time. This shows that {MV label aggregation is not representative of most individuals' line of thoughts} but only of a sentence-level common opinion. Therefore evaluating models on this MV does not give a representative performance of the models. Additionally removing the 2.33\% of uncertain CWs does not modify the distribution which is another sign that CrowdTruth is satisfaying for our purpose. 

\section{Conclusions}
We proposed a list of specifications to reflect on before creating evaluation settings for ML applications involving subjective labels, and 
showed that current ML practices are not suited to these evaluation goals. 
We suggest that ML researchers follow the guidelines of the crowdsourcing community in the design of crowdsourcing tasks, 
and supplement them by integrating 
research methodologies from other domains 
in order to factor in the variables influencing labels.
The difficulties in and implications of instantiating the specifications pointed out future challenges in crowdsourcing research to achieve the specifications in practice, and highlighted needs for future work towards creating frameworks for the easier definition of requirements. 
Appropriate evaluation requires the exploitation of EUs and CWs' personal information, however this might be a privacy issue to consider in certain applications. 





\appendix

\bibliographystyle{aaai}
\bibliography{main}

\section{Appendix}
\subsection{Distributions in the Jigsaw dataset}
\begin{figure}[h]
  \centering
  \includegraphics[width=\linewidth]{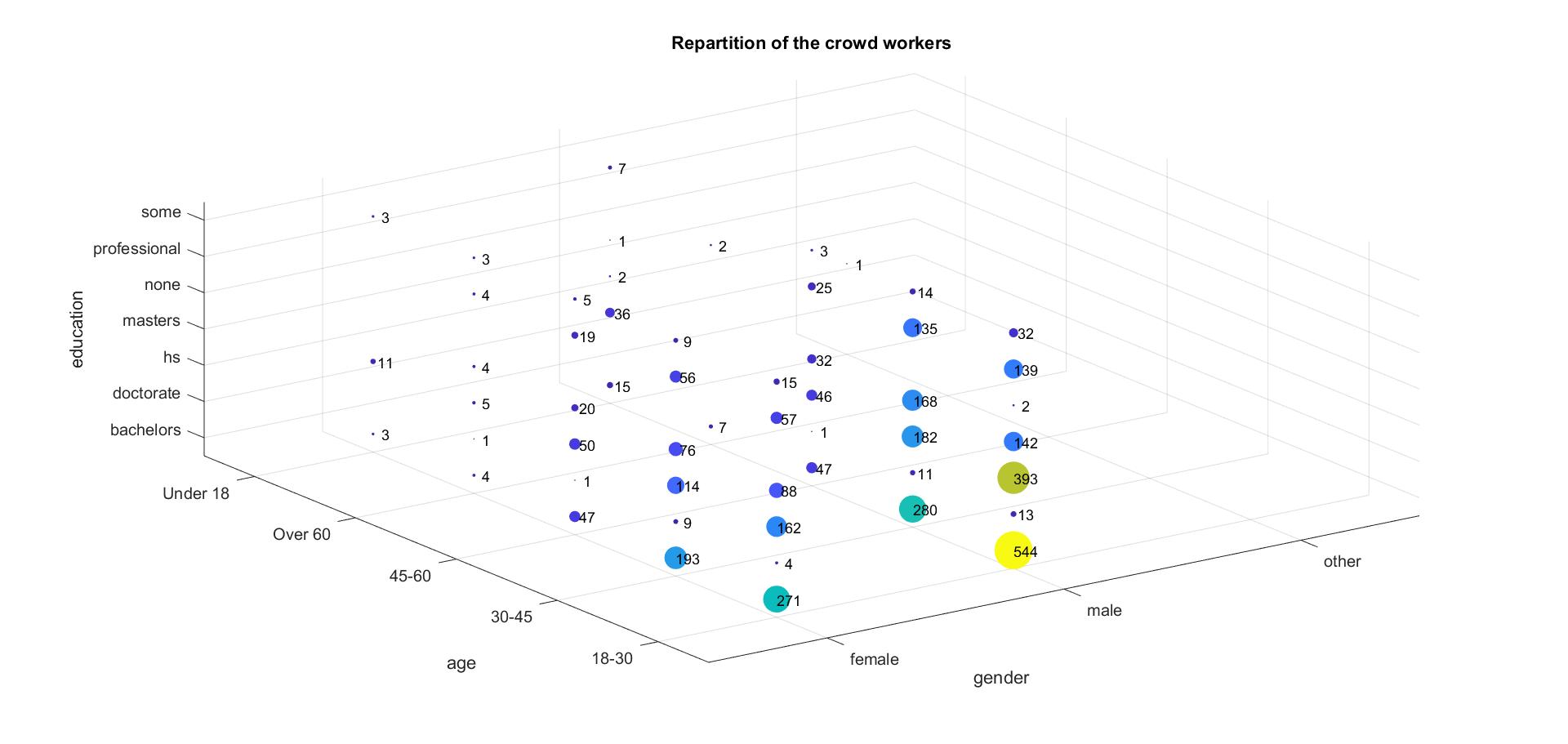}
  \caption{CWs' distribution along the protected attributes.}
 \label{fig:CW_distribution}
\end{figure}

\begin{figure}[h]
  \centering
  \includegraphics[width=0.9\linewidth]{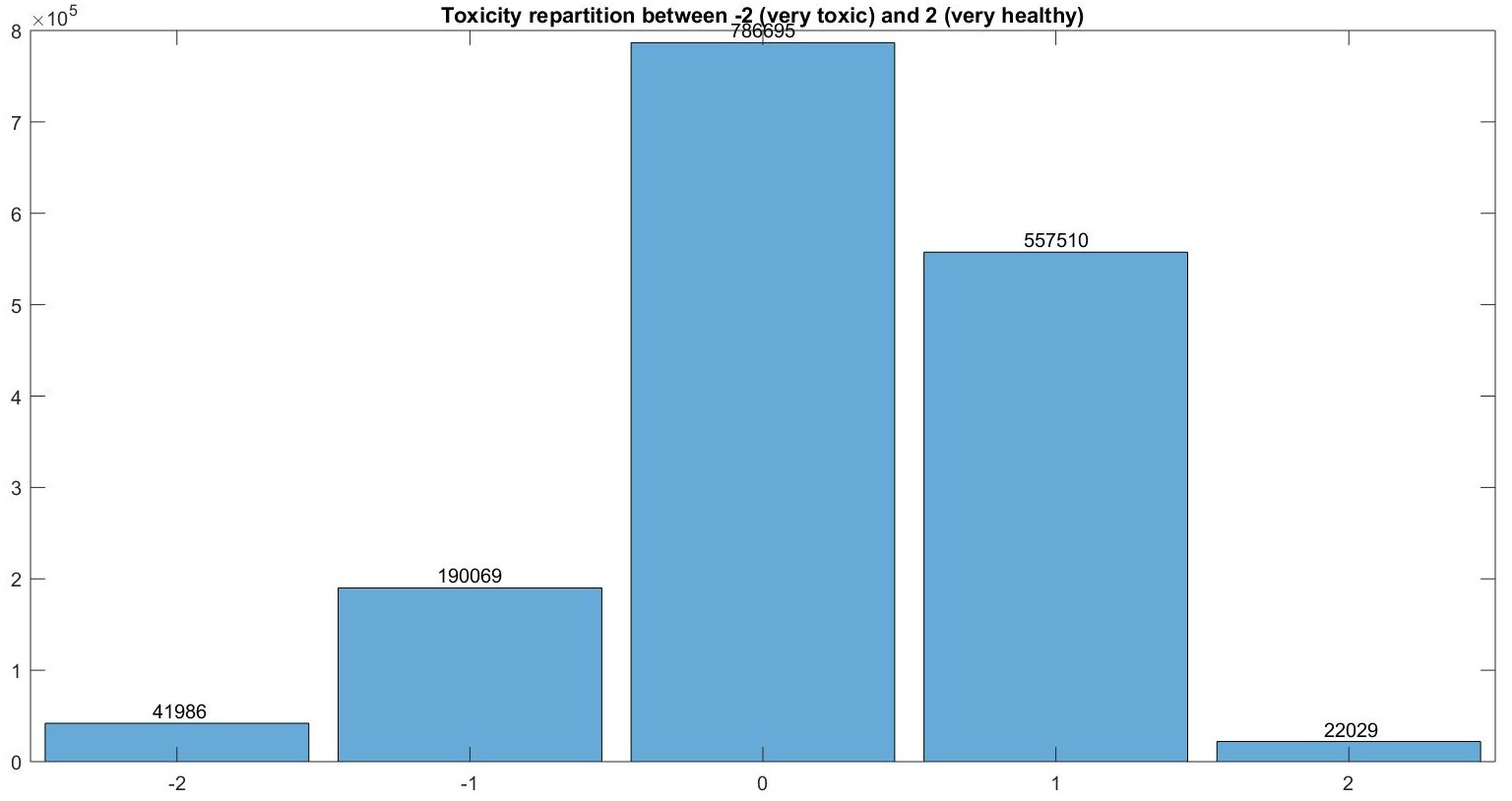}
  \caption{Distribution of collected labels.}
 \label{fig:CW_distribution1}
\end{figure}

Fig.~\ref{fig:CW_distribution} shows the distribution of $A$ in the Jigsaw dataset along the protected attributes chosen for our evaluation.
Fig.~\ref{fig:CW_distribution1} shows the distribution of the labels $L$ collected via the crowdsourcing tasks for the entire dataset.

\subsection{Metrics computed over the Jigsaw dataset}

We report metric results when grouping crowd workers into 5 bins using the full range of ADR of the dataset. The ADR-based bias measures (Tab.~\ref{tab:unfairness_models}) follow the expected trend (model 1 appears more biased than models 2 and 3) contrary to the protected attributes -based measures. 
Model 1 trained with MV labels exhibits similar performance across demographic groups because it does not differentiate them, and their distributions of ADR with the MV are similar within the training dataset (hence it makes the same amount of mistakes within each group). 
Model 2's performance is different across groups because the model distinguishes between these groups and is more accurate for groups with more training data. This is thus not an issue from the metric but from the model and the metric reflects well this issue. 
Only reporting the accuracy or the bias metrics would not have enabled to explain these trends, what is one argument for the report of the intermediate results. Additionally, if the user features were not reported and diversified in the evaluation dataset, the bias of the models could not have been detected and interpreted correctly.

\begin{table}[h]
	\footnotesize
	\centering
	\caption{Experimental results of the evaluation metrics.
	}\label{tab:unfairness_models}
	\begin{tabular}{l|ccc}
	\hline
		 & \textbf{\textit{Model 1}} & \textbf{\textit{Model 2}}  & \textbf{\textit{Model 3}} 
		 \\  \midrule[2pt]

    \multicolumn{2}{l}{First metric}  \\
    \pbox{0.08\textwidth}{\textbf{\textit{Disp.}}}& \textbf{0.94} & 0.72 & 0
		 \\ 
		\pbox{0.08\textwidth}{\textbf{\textit{Gen. perf.}}} & {0.68} & 0.63 & 1 
		 \\ \hline
		
			\multicolumn{2}{l}{Second metric}  \\
		 \pbox{0.08\textwidth}{\textbf{\textit{Disp.}}}& {0.93} & \textbf{0.96} & 0 
		 \\ 
		\pbox{0.08\textwidth}{\textbf{\textit{Gen. perf.}}} & {0.68} & 0.68 & 1 
		 \\ \hline
	\end{tabular}
\end{table}

\end{document}